\documentclass{article}

\PassOptionsToPackage{numbers, compress}{natbib}
\usepackage[pdftex]{graphicx}
\usepackage[preprint]{neurips_2026}

\usepackage[utf8]{inputenc} 
\usepackage[T1]{fontenc}    
\usepackage{hyperref}       
\usepackage{url}            
\usepackage{booktabs}       
\usepackage{amsfonts}       
\usepackage{nicefrac}       
\usepackage{microtype}      
\usepackage[table]{xcolor}
\usepackage{enumitem}
\usepackage{amsmath}
\usepackage{tcolorbox}
\usepackage{fvextra}
\usepackage{tabularx} 
\usepackage{array}   
\usepackage{fancyvrb} 
\usepackage{caption}

\title{Resolution as a First-Class Decision: Task-Conditioned Routing for \\ Efficient Multimodal Large Language Models}

\author{%
  \textbf{Zhiqiang Xia} \quad \textbf{Yang Li} \quad \textbf{Xinyuan Zhang} \quad \textbf{Yuchen Liu} \\
  \textbf{Haoyu Lu} \quad \textbf{Jiaming Xu} \quad \textbf{Runyu Shi} \quad \textbf{Ying Huang} \\
  \vspace{1.5mm}\\
  Xiaomi Corporation, Beijing, China \\
  \texttt{\{xiazhiqiang5, liyang134, zhangxinyuan8, liuyuchen8\}@xiaomi.com} \\
  \texttt{\{luhaoyu1, xujiaming1, shirunyu, huangying8\}@xiaomi.com} \\
}

\begin{document}

\maketitle


\begin{abstract}
The inference efficiency of Multimodal Large Language Models (MLLMs) is severely constrained by massive visual token sequences induced by high-resolution inputs, with computational cost scaling quadratically. Existing approaches primarily focus on downstream token compression, while overlooking a fundamental upstream inefficiency: input resolution is treated as a static, task-agnostic hyperparameter. We propose Task-Conditioned Resolution Routing (TCRR), which formulates visual compression as a task-conditioned decision and employs a lightweight cross-modal router that conditions backbone visual representations on textual semantics via feature-wise modulation and cross-attention to predict the minimal sufficient compression level per query. To support this, we curate a dataset of 500k samples across 12 task categories, labeled via a teacher–oracle pipeline to approximate Pareto-optimal compression scales. Extensive experiments across diverse architectures show that TCRR achieves a superior efficiency frontier, specifically reducing visual FLOPs by \textbf{40.9\%} and latency by \textbf{53.7\%} on Qwen3-VL-8B while preserving competitive performance. Further analysis of scaling behavior confirms that dynamically routing visual compression enables optimal resource allocation without modifying the MLLM backbone.
\end{abstract}
\section{Introduction}

\begin{figure}[t]
  \vspace*{30pt}
  \centering
  \includegraphics[width=\linewidth]{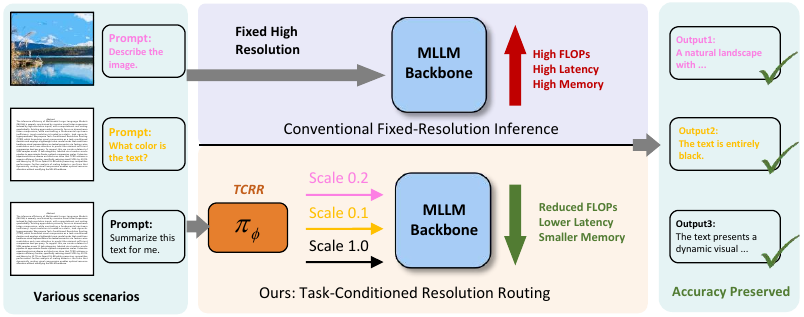}
  \captionsetup{aboveskip=18pt}
  \caption{\textbf{Task-Conditioned Resolution Routing.} By adapting resolution to textual intent, TCRR slashes computational overhead without sacrificing accuracy.}
  \label{overview}
\end{figure}

Multimodal Large Language Models (MLLMs) have demonstrated strong performance across a broad range of vision–language tasks~\cite{alayrac2022flamingo,liu2023visual}. Yet their inference efficiency remains a major bottleneck. A dominant cost arises in visual encoding: processing high-resolution inputs yields long sequences of visual tokens, which incur quadratic attention complexity and substantial latency~\cite{dosovitskiy2020image}. Prior efforts mitigate this issue through downstream token compression or pruning. While effective in reducing later-stage computation, these methods are typically applied after full-resolution encoding, leaving the most expensive part of the pipeline largely untouched~\cite{bolya2022token,zhang2024sparsevlm}.

We contend that a more fundamental inefficiency lies upstream~\cite{vasu2025fastvlm}. Current pipelines fix the input resolution as a task-agnostic hyperparameter, despite the fact that the required spatial detail is largely determined by the prompt. Optical Character Recognition (OCR), document QA, and fine-grained grounding are resolution-hungry, whereas captioning and many Visual Question Answering (VQA) queries tolerate substantial downsampling. Therefore, image-only or heuristic resolution strategies are structurally insufficient for MLLMs because they ignore prompt-dependent information demands. This leads to a simple principle: resolution should be a first-class, task-conditioned decision~\cite{rao2021dynamicvit}. Because resolution choice is upstream and irreversible, it directly controls visual token length and dominates compute/latency. Adapting it per instance yields a better efficiency–accuracy trade-off without modifying the frozen MLLM backbone.

To this end, we propose \textbf{Task-Conditioned Resolution Routing (TCRR)}, a lightweight framework that dynamically selects minimal sufficient resolution conditioned on image content and textual semantics, as shown in Fig.~\ref{overview}. Given an image--text pair, TCRR predicts a resolution scale from a predefined set using a cross-modal router that integrates feature-wise modulation and text-to-image attention. In essence, this transforms resolution selection from a static heuristic into a dynamic, task-conditioned routing process.

Training such a router presents a nontrivial supervision challenge: optimal resolution is inherently ambiguous, and multiple scales may yield equivalent task performance. To address this, we construct \textbf{Res-500k}, a task-conditioned resolution dataset comprising 500k image--text pairs across 12 task categories. Each sample is annotated via a teacher--oracle pipeline that selects the highest compression ratio while preserving semantic equivalence under GPT-based evaluation~\cite{achiam2023gpt,zheng2023judging}. This strategy provides a closer approximation of real-world utility than conventional teacher-forcing losses, and yields negligible performance degradation on held-out test sets~\cite{ouyang2022training,rafailov2023direct}.

In summary, our core contributions are threefold:
\begin{itemize}[leftmargin=*, topsep=0pt, itemsep=2pt, parsep=0pt]
    \item \textbf{Architectural Paradigm Shift:} We propose TCRR, a lightweight, plug-and-play module that formalizes resolution selection as a task-conditioned routing problem. By employing a progressive cross-modal fusion mechanism upstream of the frozen MLLM backbone, TCRR strictly outperforms downstream token merging baselines while incurring negligible routing overhead ($\sim$18ms).
    \item \textbf{Data, Supervision, and Generalization:} We curate \textbf{Res-500k}, the first large-scale dataset annotated specifically for semantic-preserving visual compression via a principled teacher--oracle pipeline. This rigorous supervision yields a routing policy that not only excels on general benchmarks but also demonstrates robust zero-shot generalization to specialized domains (e.g., medical and remote sensing).
    \item \textbf{State-of-the-Art Efficiency \& Scaling Insights:} Extensive evaluations establish a superior Pareto frontier. On Qwen3-VL-8B~\cite{qwen3-vl}, TCRR slashes visual FLOPs by \textbf{40.9\%} and system latency by \textbf{53.7\%} with negligible capability loss ($\sim$0.6\%), maintaining extreme scalability with a $43\%$ latency reduction even on the massive 235B architecture. Crucially, we reveal that adaptive resolution synergy depends heavily on tokenization granularity: continuous patch-based architectures unlock significantly higher efficiency gains compared to rigid tiling designs, providing a vital design principle for next-generation efficient MLLMs.
\end{itemize}
    
\section{Related Work}

\paragraph{Visual Compression vs. Upstream Resolution.}
Methods mitigating MLLM visual costs broadly divide into \textit{downstream} token reduction and \textit{upstream} resolution scaling. Downstream approaches~\cite{bolya2022token,zhang2024sparsevlm,marin2023token,yang2025visionzip} reduce later-stage computation but operate \textit{after} full-resolution encoding, failing to alleviate the Vision Transformer's (ViT) peak memory and upfront computational tax. Conversely, existing upstream strategies~\cite{hypervl} circumvent the ViT bottleneck but remain \textit{task-agnostic}, relying purely on image heuristics. TCRR bridges this gap by conditioning the resolution decision directly on the textual prompt, dynamically aligning spatial resource allocation with the query's semantic demands.

\paragraph{Conditional Computation \& Routing.}
Adaptive paradigms like Mixture-of-Experts~\cite{shazeer2017outrageously,fedus2022switch} or layer skipping~\cite{wang2018skipnet,schuster2022confident} optimize computation \textit{within} the network. However, these intra-model approaches assume fixed-fidelity inputs and often demand invasive architectural modifications or joint pre-training. TCRR instead shifts the decision boundary to the \textit{input level}. By formalizing resolution selection as a task-conditioned routing problem, TCRR acts as a lightweight, plug-and-play pre-filter. This governs sequence length at the source, optimizing the efficiency-accuracy trade-off while keeping the heavy MLLM backbone strictly frozen.    
\section{Problem Formulation}
\label{sec:problem}

We formally define resolution selection within efficiency-constrained MLLMs. Let $\mathcal{M}_{\Theta}$ denote an MLLM processing a visual input $I_r$ (resampled to resolution $r = h \times w$) and a textual prompt $T$. Since the visual encoder's self-attention exhibits quadratic complexity, the computational cost scales quadratically with resolution: $\mathcal{C}(r) \propto (r/P^2)^2 \propto r^2$, where $P$ is the patch size. While prior dynamic-resolution methods mitigate this, they are typically solely image-conditioned, ignoring the diverse spatial demands imposed by different textual instructions.

We posit that optimal resolution is intrinsically task-dependent. For any instance $(I, T)$, we seek a minimal resolution $r^* \in \mathcal{S}$ from a discrete set of candidates that minimizes cost without degrading the reference high-resolution response $Y_{\text{high}}$:
\begin{equation}
\label{eq:objective}
    r^* = \operatorname*{argmin}_{r \in \mathcal{S}} \mathcal{C}(r) \quad \text{s.t.} \quad \mathcal{L}(Y_r, Y_{\text{gt}}) \le \mathcal{L}(Y_{\text{high}}, Y_{\text{gt}}) + \delta,
\end{equation}
where $\delta$ is a negligible tolerance margin. Solving Eq.~\ref{eq:objective} via exhaustive inference is intractable. Therefore, we approximate this oracle decision via a lightweight, task-conditioned routing policy $\pi_{\phi}$. Operating on a fixed-resolution proxy image $I_{\text{fix}}$, the router predicts a continuous scaling factor $\tilde{s} \leftarrow \pi_{\phi}(\text{Resize}(I, r_{\text{fix}}), T)$. The frozen backbone then processes the image at $\hat{r} = \tilde{s} \cdot r_{\text{fix}}$. By ensuring the router's overhead is negligible (typically $<5\%$ of the backbone cost), reductions in $r$ translate directly to strict wall-clock acceleration.

\section{Dataset Construction: Res-500k}
\label{sec:dataset}

Since intrinsic optimal resolutions are latent, we construct \textbf{Res-500k}, a large-scale dataset annotated via a rigorous \textit{Teacher--Oracle} pipeline. A naive approach to label resolution sufficiency is to monitor the MLLM's next-token prediction loss (teacher-forcing). However, this is a fundamentally flawed proxy: a low loss might simply reflect strong language priors rather than sufficient visual clarity, while a high loss could stem from benign phrasing differences. To bypass this, our pipeline directly optimizes for \textit{semantic equivalence}.

\subsection{The Teacher--Oracle Pipeline \& Quality Assurance}

Our pipeline determines the minimal sufficient scale $s^*$ for each instance $(I, T)$. We first generate 10 candidate resolutions using uniform area scale factors $\mathcal{S} = \{0.1, 0.2, \dots, 1.0\}$. Using Qwen3-VL-8B as the teacher, we obtain a reference response $Y_{\text{ref}}$ at native resolution ($I_{1.0}$), followed by hypothesis responses $\{Y_s\}_{s \in \mathcal{S}}$ across all candidate scales.

\textbf{Rigorous Quality Assurance.} To strictly evaluate semantic equivalence $\mathcal{J}(Y_s, Y_{\text{ref}} \mid T) \in \{0, 1\}$, we conducted comprehensive pilot studies across multiple LLM judges (Qwen-235B-A22B, GPT-4o, GPT-5.1). We selected GPT-5.1 as it achieved flawless semantic alignment. (The exact prompt utilized for this oracle judgment is provided in Appendix~\ref{app:prompt}.) Crucially, when applying this oracle selection strategy ($s^*$ where $\mathcal{J}=1$) to the 12 evaluation benchmarks, it preserved exactly 100\% of the uncompressed performance (establishing our Oracle Bound). To guarantee absolute reliability, human experts (Master's degree or higher) manually evaluated 1,200 randomly sampled instances (100 per category). This verification revealed a 100\% alignment with GPT-5.1's equivalence judgments, confirming the absence of systemic bias and strictly preventing data leakage from evaluation sets.

\textbf{Target Rescaling.} Since our router operates on a fixed proxy size $r_{\text{fix}} \ge 384^2$, predicting $s^*$ directly is ambiguous across varied original image dimensions $r_{\text{orig}}$. Following HyperVL~\cite{hypervl}, we rescale the regression target to a resolution-agnostic multiplier: $\tilde{s}^* = s^* \cdot (r_{\text{orig}} / r_{\text{fix}})$, decoupling the prediction from raw input dimensions.

\subsection{Dataset Statistics and Diversity}

Res-500k comprises 500,000 instances balanced across 12 distinct task categories (Table~\ref{tab:dataset}). To ensure alignment with state-of-the-art training protocols, task sampling ratios are calibrated against leading multimodal datasets~\cite{gu2024infinity,wiedmann2025finevision,an2025llava}. It covers General Vision-Language (42\%, mostly low/mixed resolution demands), Document \& Information Extraction (37\%, high demands), and GUI \& Agent Navigation (21\%, very high demands). A stratified hold-out test set of 12,000 samples guarantees unbiased evaluation.

\begin{table}[t]
\centering
\caption{Task distribution and resolution in \textbf{Res-500k}. The dataset is explicitly balanced to cover diverse resolution priors, ranging from semantic gist (Low) to pixel-level precision (Very High).}
\label{tab:dataset}
\small 
\begin{tabular}{l c l}
\toprule
\textbf{Task Category} & \textbf{Proportion} & \textbf{Resolution Demand \& Key Factor} \\
\midrule
\multicolumn{3}{l}{\textit{\textbf{General Vision-Language}}} \\
VQA & 22\% & \textbf{Mixed} (Context-dependent complexity) \\
Captioning & 10\% & \textbf{Low} (Global semantic gist) \\
Grounding & 5\% & \textbf{Medium} (Bounding box localization) \\
Counting & 5\% & \textbf{High} (Object separation) \\
\midrule
\multicolumn{3}{l}{\textit{\textbf{Document \& Text}}} \\
OCR & 15\% & \textbf{High} (Character legibility) \\
STEM QA & 12\% & \textbf{High} (Symbol \& diagram parsing) \\
Doc Understanding & 5\% & \textbf{High} (Dense text reading) \\
Chart QA & 5\% & \textbf{Medium-High} (Trend \& value analysis) \\
\midrule
\multicolumn{3}{l}{\textit{\textbf{GUI \& Agent}}} \\
GUI Grounding & 8\% & \textbf{Very High} (Pixel coordinate precision) \\
GUI VQA & 7\% & \textbf{High} (Small icon/text recognition) \\
GUI Caption & 5\% & \textbf{Medium} (Holistic screen description) \\
GUI Action & 1\% & \textbf{Mixed} (Element interaction) \\
\bottomrule
\end{tabular}
\end{table}

\section{Task-Conditioned Resolution Routing}

We propose TCRR, a plug-and-play module that transforms the static inference paradigm of MLLMs into an adaptive, instance-aware decision process. Acting as a semantic pre-filter upstream of the \textbf{unmodified} MLLM backbone, TCRR predicts the minimal sufficient resolution scale $\tilde{s}$ by explicitly modeling the interplay between visual semantics and textual intent.

\subsection{Inference Pipeline Overview}
\begin{figure*}[t]
  \vspace*{10pt}
  \centering
  \includegraphics[width=0.95\textwidth]{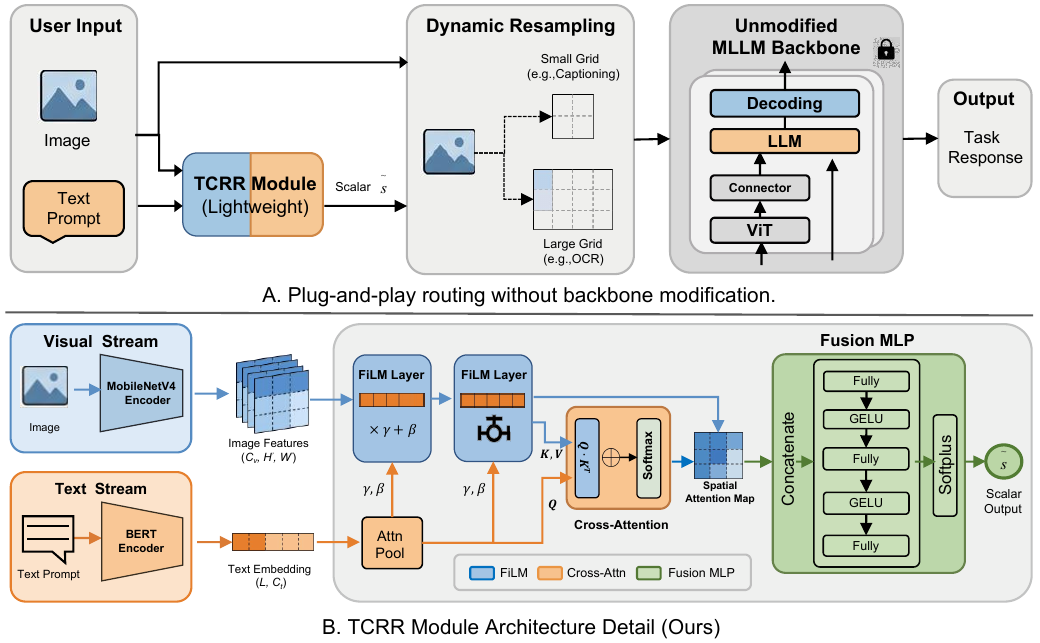}
  \caption{\textbf{The architecture of TCRR.} The lightweight router (\textbf{bottom}) fuses visual features and the text prompt via cross-attention to predict an optimal continuous scale factor, dynamically resizing the raw image before it enters the \textbf{unmodified} MLLM backbone (\textbf{top}).}
  \label{fig:overview}
\end{figure*}

As illustrated in Figure~\ref{fig:overview}, the inference trajectory is bifurcated. The \textbf{TCRR} module first processes the raw input pair $(I, T)$ through a lightweight architecture to output a continuous scale factor $\tilde{s}$. This prediction governs the \textbf{Dynamic Resampling} module, where the original image is interpolated to the target resolution:
\begin{equation}
    \hat{r} = \tilde{s} \cdot r_{\text{fix}}, \quad \tilde{I} = \text{Resize}(I, \hat{r}).
\end{equation}
The resampled image $\tilde{I}$ and the original prompt $T$ are subsequently routed to the \textbf{unmodified} MLLM backbone. Crucially, TCRR introduces zero learnable parameters to the backbone, guaranteeing seamless integration with existing models while incurring strictly negligible computational overhead.

\subsection{The Architecture of TCRR}
\label{sec:routernet}

To balance computational efficiency with cross-modal expressiveness, TCRR employs a dual-encoder architecture coupled with a \textit{Progressive Fusion} mechanism. This mechanism systematically aggregates task-conditioned features via sequential feature-wise modulation and spatial cross-attention.

\subsubsection{Unimodal Feature Extraction}

\paragraph{Visual Encoder.}
We adopt MobileNetV4~\cite{mobilenet-v4} as the visual proxy backbone. The encoder processes a fixed-resolution downsampled image ($I_{\text{fix}} \in \mathbb{R}^{384 \times 384 \times 3}$), outputting a hierarchical feature pyramid. We extract the dense feature map $\mathbf{V} \in \mathbb{R}^{C_v \times H' \times W'}$ from the final convolutional stage to serve as the foundational visual representation.

\paragraph{Text Encoder.}
To encode the instructional prompt $T$, we utilize BERT-Tiny~\cite{berttiny}. Rather than relying solely on the pooled \texttt{[CLS]} token—which may dilute fine-grained instructional cues—we apply a learnable \textit{Attention Pooling} layer over the full sequence of hidden states $\mathbf{T}_{\text{seq}}$ to derive a task-specific global textual embedding $\mathbf{t}_{\text{global}} \in \mathbb{R}^{C_t}$:
\begin{equation}
    \mathbf{t}_{\text{global}} = \sum_{i=1}^{L} \alpha_i \mathbf{T}_{\text{seq}}^{(i)}, \quad \alpha = \text{Softmax}(\mathbf{T}_{\text{seq}} W_{\text{attn}}).
\end{equation}
This mechanism enables the router to dynamically attend to resolution-critical keywords (e.g., ``read the small text", ``count the objects") within the prompt.

\subsubsection{Task-Conditioned Cross-Modal Interaction}

Determining the optimal resolution necessitates answering two questions: \textit{what} visual information is semantically required, and \textit{where} is it spatially located? We model this through two complementary, sequentially applied operations.

\paragraph{Global Task Modulation via FiLM.}
Global semantic intent dictates macroscopic detail. We inject this prior into the visual space using Feature-wise Linear Modulation (FiLM)~\cite{perez2018film}. Affine parameters are predicted from $\mathbf{t}_{\text{global}}$ to channel-wise modulate $\mathbf{V}$:
\begin{equation}
    [\gamma, \beta] = \text{MLP}_{\text{FiLM}}(\mathbf{t}_{\text{global}}), \quad \mathbf{V}_{\text{mod}} = \gamma \odot \mathbf{V} + \beta.
\end{equation}
This acts as a semantic filter, amplifying task-relevant channels while attenuating background noise.

\paragraph{Spatial Awareness via Cross-Attention.}
To capture localized, spatially varying resolution demands, we employ a lightweight Cross-Attention layer. Here, the global text embedding acts as the query to interrogate the modulated visual feature map. The attention mechanism aggregates visual cues strictly weighted by their relevance to the instructional text:
\begin{align}
    Q &= W_q \mathbf{t}_{\text{global}}, \nonumber \\
    K, V &= W_k \text{Flatten}(\mathbf{V}_{\text{mod}}), W_v \text{Flatten}(\mathbf{V}_{\text{mod}}), \nonumber \\
    \mathbf{z}_{\text{cross}} &= \text{Attention}(Q, K, V).
\end{align}
This produces a spatially-aware descriptor $\mathbf{z}_{\text{cross}}$ that explicitly encodes the visual complexity of the image regions most critical to fulfilling the query.

\subsubsection{Decision Head and Training Objective}


To form the final decision, we derive a global visual context vector $\mathbf{z}_{\text{img}} = \text{AttnPool}(\mathbf{V}_{\text{mod}})$. These multi-level signals are fused and projected to a strictly positive scalar via a Softplus activation:
\begin{equation}
    \tilde{s} = \text{Softplus}(\text{MLP}_{\text{head}}(\text{Concat}([\mathbf{z}_{\text{img}} + \mathbf{z}_{\text{cross}}, \mathbf{t}_{\text{global}}]))).
\end{equation}

\paragraph{Optimization.}
The router is trained end-to-end to minimize the mean squared error (MSE) against the rescaled ground-truth targets derived from our Teacher-Oracle pipeline (Section~\ref{sec:dataset}):
\begin{equation}
    \mathcal{L}_{\text{router}} = \lVert \tilde{s} - \tilde{s}^* \rVert_2^2.
\end{equation}
This continuous regression objective forces the router to learn a smooth mapping from the multimodal input space to the optimal relative compression scale. A detailed computational complexity analysis confirming TCRR's negligible overhead is provided in Appendix~\ref{app:complexity}.    
\section{Experiments}
\label{sec:experiments}

We rigorously evaluate TCRR to validate its impact on the efficiency--accuracy Pareto frontier. To comprehensively demonstrate our contributions, our empirical analysis is structured around five core research questions:
\begin{itemize}[leftmargin=*, topsep=0pt, itemsep=2pt, parsep=0pt]
    \item \textbf{RQ1 (Efficiency \& Baselines):} Can TCRR break the upfront computational bottleneck of MLLM visual encoders and fundamentally outperform downstream token merging paradigms?
    \item \textbf{RQ2 (Ablation \& Architecture):} How do the individual cross-modal components and the progressive fusion design drive both routing accuracy and low-overhead efficiency?
    \item \textbf{RQ3 (Generalization):} Do the routing gains generalize to unseen specialized domains (e.g., medical, remote sensing) and across distinct dynamic resolution implementations (e.g., continuous patch scaling vs. discrete sub-image tiling)?
    \item \textbf{RQ4 (Scaling \& Granularity):} How do the training data regime, action space granularity, and router model capacity dictate the efficiency--accuracy trade-off?
    \item \textbf{RQ5 (Oracle Gap):} How closely does TCRR's routing policy approximate the theoretical optimal compression limit under strict semantic equivalence?
\end{itemize}

\subsection{Experimental Setup}

\textbf{Benchmarks and Task Coverage.}
We evaluate TCRR on a comprehensive suite of 12 primary multimodal benchmarks, encompassing MMBench V1.1, MMStar, MMMU, MathVista, HallusionBench, OCRBench, MMVet, TextVQA, ChartQA, DocVQA, InfoVQA, and MME-RealWorld-CN~\cite{mmbench,mmstar,mmmu,mathvista,hallusionbench,ocrbench,mmvet,textvqa,chartqa,docvqa,infographicvqa,mmerealworld}. We deliberately curate a mix of text-intensive benchmarks and high-resolution scenarios to strictly evaluate tasks requiring fine-grained visual details against those robust to aggressive downsampling. Furthermore, to rigorously test zero-shot out-of-domain generalization, we include OmniMedVQA (Medical) and LRS-VQA (Remote Sensing).

\textbf{Backbones and Baselines.}
Experiments are conducted across diverse MLLM families: \textbf{Qwen3-VL} (2B, 4B, 8B, 30B-A3B, 32B, 235B-A22B) and \textbf{InternVL3.5} (1B--38B)~\cite{qwen3-vl,internvl}. All backbones remain strictly unmodified; TCRR is inserted as a plug-and-play upstream module, ensuring architectural fairness. We also benchmark against VisionZip~\cite{yang2025visionzip}, a state-of-the-art token merging technique, to directly compare our upstream resolution routing against downstream token dropping.

\textbf{Metrics and Platform.}
We report accuracy, Mean Absolute Error (MAE) for scale prediction, FLOPs per sample (in TeraFLOPs, T), and visual token reduction. \textbf{System Latency} is rigorously defined as the sum of the visual encoder, projector, and LLM prefill stages (explicitly including the router's overhead) in milliseconds (ms), excluding autoregressive decoding variance. Measurements are deterministic, executed on NVIDIA H20 GPUs. The default TCRR employs MobileNetV4-Medium~\cite{mobilenet-v4} and BERT-Tiny~\cite{berttiny}, trained on Res-500k (details in Appendix~\ref{app:implementation}).

\begin{table*}[t]
\caption{\textbf{Main Results on Qwen3-VL Family.} TCRR achieves substantial reductions in FLOPs and latency while preserving average accuracy across 12 multimodal benchmarks. \textbf{Abbreviations:} MMB (MMBench), MMS (MMStar), MathV (MathVista), Hallu. (HallusionBench), TVQA (TextVQA), ChQA (ChartQA), DVQA (DocVQA), IVQA (InfoVQA), MRWC (MME-RealWorld-CN).}
\label{tab:qwen3_tcrr_full}
\centering
\scriptsize 
\setlength{\tabcolsep}{0pt} 
\begin{tabular*}{\textwidth}{@{\extracolsep{\fill}} l cccccccccccc ccc @{}}
\toprule
Model 
& MMB & MMS & MMMU & MathV & Hallu. & OCR & MMVet & TVQA & ChQA & DVQA & IVQA & MRWC & Avg. & FLOPs (T) & Lat. (ms) \\
\midrule
2B & 76.8 & 54.7 & 46.3 & 51.1 & 47.1 & 84.0 & 40.7 & 79.6 & 78.0 & 92.5 & 72.2 & 55.2 & 64.9 & 8.5 & 524.1 \\
\quad +TCRR & 76.8 & 54.7 & 45.7 & 51.4 & 46.4 & 84.0 & 40.5 & 79.7 & 77.5 & 92.2 & 69.4 & 53.3 & 64.3 & \textbf{4.8}{\tiny(-43\%)} & \textbf{236.9}{\tiny(-55\%)} \\
\midrule
4B & 82.9 & 62.3 & 55.0 & 64.2 & 55.1 & 84.8 & 49.7 & 81.6 & 81.8 & 94.6 & 79.4 & 62.7 & 71.2 & 14.7 & 672.0 \\
\quad +TCRR & 82.6 & 62.7 & 54.9 & 64.5 & 54.2 & 84.9 & 49.9 & 81.6 & 81.3 & 94.8 & 77.9 & 60.9 & 70.8 & \textbf{8.5}{\tiny(-42\%)} & \textbf{319.8}{\tiny(-52\%)} \\
\midrule
8B & 85.0 & 65.3 & 57.4 & 65.6 & 58.0 & 87.0 & 52.9 & 83.3 & 82.8 & 95.6 & 83.1 & 64.5 & 73.4 & 23.7 & 1037.7 \\
\quad +TCRR & 84.9 & 65.4 & 54.0 & 65.7 & 57.5 & 87.1 & 53.2 & 83.4 & 82.7 & 95.4 & 81.3 & 62.6 & 72.8 & \textbf{14.0}{\tiny(-41\%)} & \textbf{480.8}{\tiny(-54\%)} \\
\midrule
30B-A3B & 86.2 & 68.1 & 60.6 & 73.3 & 60.5 & 89.4 & 49.9 & 84.5 & 86.2 & 95.4 & 82.9 & 66.0 & 75.2 & 12.8 & 2486.2 \\
\quad +TCRR & 86.3 & 68.9 & 56.4 & 73.6 & 60.9 & 88.8 & 50.8 & 84.8 & 86.0 & 95.2 & 80.4 & 64.1 & 74.7 & \textbf{7.2}{\tiny(-44\%)} & \textbf{1313.0}{\tiny(-47\%)} \\
\midrule
32B & 87.8 & 72.1 & 61.8 & 73.9 & 60.0 & 89.0 & 63.0 & 83.1 & 84.2 & 96.0 & 88.0 & 69.7 & 77.4 & 78.2 & 2383.9 \\
\quad +TCRR & 87.8 & 71.9 & 57.3 & 73.8 & 60.8 & 88.8 & 61.4 & 83.2 & 83.4 & 96.1 & 85.0 & 67.6 & 76.4 & \textbf{45.6}{\tiny(-42\%)} & \textbf{1235.1}{\tiny(-48\%)} \\
\midrule
235B-A22B & 90.6 & 73.7 & 65.2 & 76.4 & 62.9 & 91.4 & 59.5 & 86.9 & 88.9 & 96.4 & 88.2 & 69.0 & 79.1 & 54.5 & 9365.9 \\
\quad +TCRR & 90.5 & 74.2 & 62.3 & 76.7 & 62.5 & 91.5 & 57.2 & 86.7 & 88.6 & 96.2 & 87.0 & 67.2 & 78.4 & \textbf{32.7}{\tiny(-40\%)} & \textbf{5329.4}{\tiny(-43\%)} \\
\bottomrule
\end{tabular*}
\end{table*}

\subsection{Main Results \& Baselines (RQ1)}

\textbf{Breaking the Efficiency Barrier.} Table~\ref{tab:qwen3_tcrr_full} establishes TCRR's superiority across the Qwen3-VL family. On Qwen3-VL-8B, TCRR slashes FLOPs by \textbf{40.9\%} and System Latency by \textbf{53.7\%}. This massive compute reduction is functionally lossless, with average capability dropping by merely \textbf{0.61\%}. Notably, on OCRBench, accuracy slightly improves, indicating that dynamic resolution effectively filters out redundant visual noise. These gains are scalable, saving over 1.1 seconds per query on the 32B model, and maintaining a $43\%$ latency reduction on the massive 235B-A22B architecture.

\textbf{Superiority Over Token Merging.} Table~\ref{tab:visionzip_comparison} juxtaposes TCRR against VisionZip, a leading post-hoc token merging technique. Upstream routing proves strictly superior: at a $\sim 45-50\%$ token reduction regime, TCRR retains $99.2\%$ of capability, whereas VisionZip degrades to $96.6\%$. This empirically validates our core hypothesis: discarding spatial structures \textit{after} expensive high-resolution ViT encoding is vastly inferior to computing representations at an optimal, natively lower resolution.

\begin{table}[t]
\centering
\scriptsize 
\setlength{\tabcolsep}{2.5pt} 
\begin{minipage}[t]{0.48\textwidth}
\vspace{0pt} 
\centering
\caption{\textbf{Comparison with Token Merging.} TCRR outperforms VisionZip on Qwen3-VL-8B at equivalent compression regimes.}
\label{tab:visionzip_comparison}
\begin{tabular}{l ccc}
\toprule
\textbf{Method} & \textbf{Avg. Acc.} & \textbf{Acc. Retained} & \textbf{Token Red.} \\
\midrule
Baseline (8B) & 73.4 & 100\% & - \\
\midrule
VisionZip ($\text{ratio}=0.8$) & 72.6 & 98.9\% & 20.0\% \\
VisionZip ($\text{ratio}=0.75$) & 72.3 & 98.5\% & 25.0\% \\
VisionZip ($\text{ratio}=0.5$) & 70.9 & 96.6\% & 50.0\% \\
VisionZip ($\text{ratio}=0.33$) & 67.7 & 92.2\% & 66.7\% \\
\midrule
\textbf{+ TCRR (Ours)} & \textbf{72.8} & \textbf{99.2\%} & \textbf{45.8\%} \\
\bottomrule
\end{tabular}
\end{minipage}\hfill
\begin{minipage}[t]{0.48\textwidth}
\vspace{0pt} 
\centering
\caption{\textbf{Component Ablation on Qwen3-VL-8B.} Text conditioning is the main driver, while FiLM and Cross-Attention fine-tune the decision boundary.}
\label{tab:routernet_ablation}
\begin{tabular}{l cc ccc}
\toprule
\textbf{Config.} & \textbf{Acc.} & \textbf{MAE} & \textbf{FLOPs} & \textbf{Lat.} & \textbf{Token Red.} \\
\midrule
Vision-Only & 65.9 & 1.42 & \textbf{9.8} & \textbf{321.8} & \textbf{85.7\%} \\
+ Text Concat & 70.0 & 1.27 & 12.9 & 425.7 & 56.1\% \\
+ Cross-Attn & 70.8 & 1.22 & 15.0 & 483.4 & 40.4\% \\
+ FiLM & 71.3 & 1.23 & 15.0 & 484.6 & 40.8\% \\
\textbf{Full TCRR} & \textbf{72.8} & \textbf{1.14} & 14.0 & 480.8 & 45.8\% \\
\bottomrule
\end{tabular}
\end{minipage}
\end{table}

\subsection{Ablation Studies (RQ2)}

\textbf{Component Analysis.} Table~\ref{tab:routernet_ablation} systematically deconstructs TCRR. We observe that text conditioning is absolutely critical: adding basic text concatenation yields a $+4.0\%$ accuracy surge over the vision-only baseline, proving that minimal sufficient resolution is inherently query-dependent. The integration of Cross-Attention (+0.8\%) and FiLM (+0.5\%) sequentially refines precision, yielding the lowest MAE (1.14), validating our strategy of fusing global semantics with spatial demands.

\textbf{Architecture Comparison.} Beyond individual components, the structural paradigm of fusion is paramount. While multi-stage, interleaved cross-attention models are common, our empirical analysis reveals that our lightweight, \textbf{single-stage fusion achieves the optimal balance of accuracy and overhead}. Specifically, compared to a multi-stage counterpart, our single-stage TCRR achieves matching accuracy ($72.8\%$ vs $72.7\%$) while executing significantly faster ($18.1$ms vs $22.8$ms router latency), strictly adhering to the requirement that routing overhead remains negligible (detailed architecture comparison is provided in Appendix~\ref{app:fusion_architecture}).

\begin{table}[t]
\centering
\scriptsize 
\setlength{\tabcolsep}{2.5pt} 
\begin{minipage}[t]{0.48\textwidth} 
\vspace{0pt} 
\centering
\caption{\textbf{Specialized Domain Generalization.} TCRR demonstrates robust zero-shot generalization on Medical and Remote Sensing datasets.}
\label{tab:domain_generalization}
\begin{tabular}{l cccc} 
\toprule
\textbf{Benchmark} & \textbf{Qwen3-VL-8B} & \textbf{+TCRR} & \textbf{Acc. Ret.} & \textbf{Token Red.} \\
\midrule
OmniMedVQA & 81.3 & 82.1 & 100\% & 34.2\% \\
LRS-VQA & 34.8 & 33.2 & 95.4\% & 71.2\% \\
\bottomrule
\end{tabular}
\end{minipage}\hfill
\begin{minipage}[t]{0.48\textwidth} 
\vspace{0pt} 
\centering
\caption{\textbf{Cross-Family Analysis.} Qwen3-VL's flexible tokenization enables greater savings than InternVL's rigid tiling.}
\label{tab:cross_family}
\begin{tabular}{l cccc}
\toprule
\textbf{Family (Avg.)} & \textbf{Acc. Ret.} & \textbf{FLOPs Red.} & \textbf{Lat. Red.} & \textbf{Token Red.} \\
\midrule
Qwen3-VL & \textbf{99.2\%} & \textbf{41.6\%} & \textbf{48.4\%} & \textbf{45.8\%} \\
InternVL3.5 & 98.0\% & 16.3\% & 11.9\% & 9.7\% \\
\bottomrule
\end{tabular}
\end{minipage}
\end{table}

\subsection{Generalization and Tokenization (RQ3)}

\textbf{Out-of-Domain Generalization.} As demonstrated in Table~\ref{tab:domain_generalization}, TCRR exhibits robust zero-shot generalization to highly specialized distributions without requiring domain-specific fine-tuning. On OmniMedVQA, TCRR reduces tokens by $34.2\%$ while simultaneously increasing accuracy (+0.8\%), and on LRS-VQA, it preserves $95.4\%$ capability while aggressively pruning $71.2\%$ of the visual payload.

\textbf{Impact of Tokenization Granularity.} Table~\ref{tab:cross_family} highlights the impact of native architecture on routing efficacy. Qwen3-VL benefits significantly more (FLOPs $\downarrow 42\%$) than InternVL3.5 (FLOPs $\downarrow 16\%$). This disparity yields a foundational design principle: adaptive resolution is most potent when paired with continuous, fine-grained tokenization. Qwen3-VL's patch-based paradigm allows scale predictions to translate directly into token savings, whereas InternVL's fixed $\ge 448 \times 448$ tiling quantizes the action space, restricting realizable efficiency.

\begin{table}[t]
\centering
\setlength{\tabcolsep}{4pt} 
\begin{minipage}[t]{0.48\textwidth} 
\vspace{0pt} 
\centering
\scriptsize 
\caption{\textbf{Data Scaling.} Router performance improves steadily with supervision, saturating at 500k samples.}
\label{tab:training_data_scaling}
\begin{tabular}{l cc ccc}
\toprule
\textbf{Samples} & \textbf{Acc.} & \textbf{MAE} & \textbf{FLOPs} & \textbf{Lat.} & \textbf{Token Red.} \\
\midrule
100k & 72.2 & 1.37 & 10.2 & 344.1 & 61.6\% \\
200k & 72.3 & 1.22 & \textbf{9.3} & \textbf{313.9} & \textbf{65.2\%} \\
300k & 72.5 & 1.18 & 11.9 & 398.1 & 54.7\% \\
400k & 72.5 & 1.15 & 11.8 & 399.6 & 54.7\% \\
\textbf{500k} & \textbf{72.8} & \textbf{1.14} & 14.0 & 480.8 & 45.8\% \\
\bottomrule
\end{tabular}
\end{minipage}\hfill
\begin{minipage}[t]{0.48\textwidth} 
\vspace{0pt} 
\centering
\includegraphics[width=0.85\linewidth]{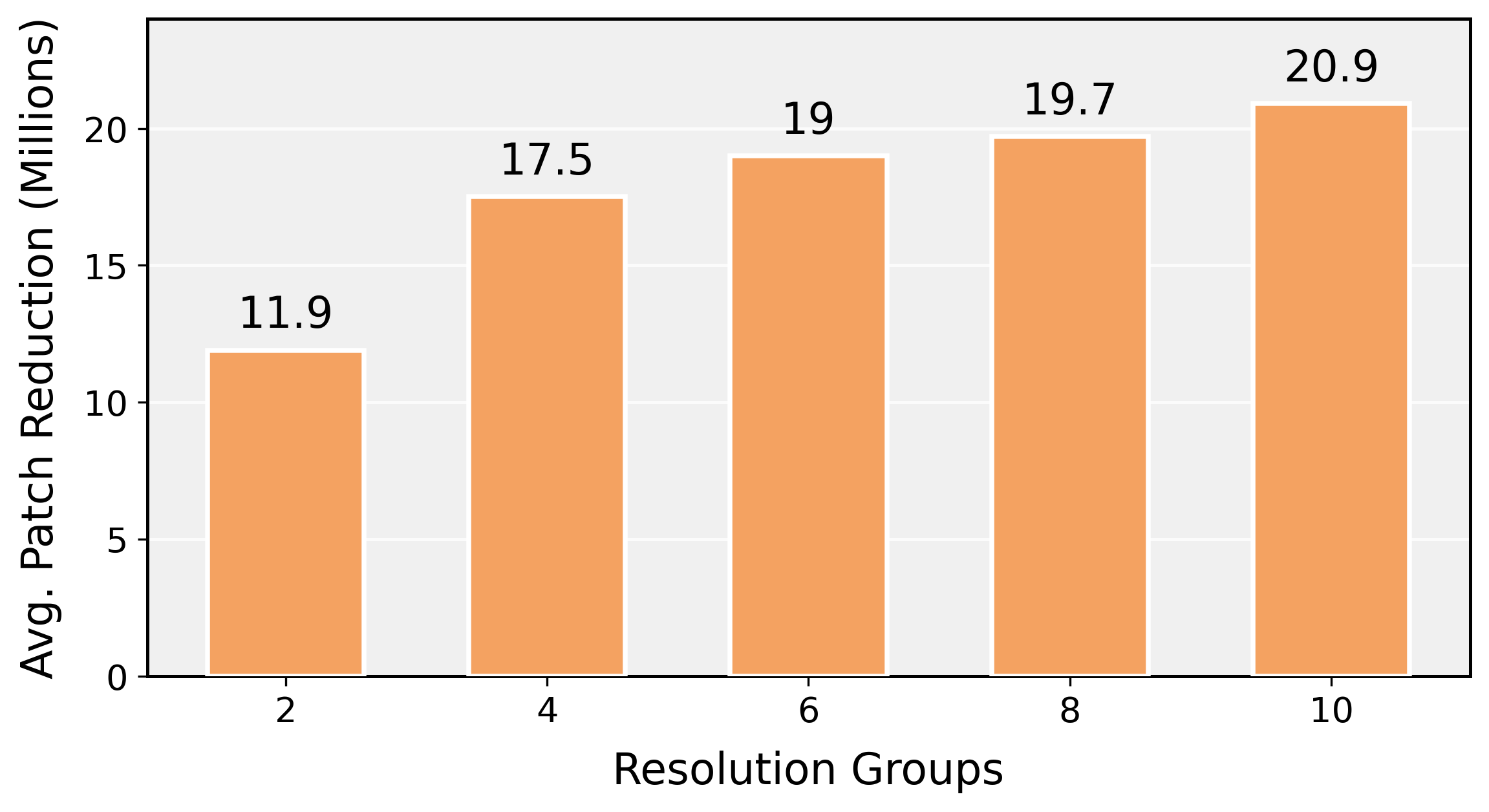} 
\captionof{figure}{\textbf{Resolution granularity vs. token reduction.} Patch reduction across 12 benchmarks.}
\label{fig:resolution_bins}
\end{minipage}
\end{table}

\subsection{Scaling Laws and Granularity (RQ4)}

\textbf{Data and Model Scaling.} Table~\ref{tab:training_data_scaling} demonstrates a log-linear MAE improvement with training data scale, confirming that \textbf{Res-500k} provides the compute-optimal saturation point for generalization. Correspondingly, Table~\ref{tab:router_model_size} sweeps 15 router configurations to map the Pareto frontier, pinpointing MobileNet-Medium + BERT-Tiny as the optimal "sweet spot". While MobileNet-Small offers minor latency advantages, its $224 \times 224$ initialization critically limits feature granularity. Conversely, scaling to Large backbones causes inference costs to balloon, undermining the core efficiency objective.

\textbf{Action Granularity.} Figure~\ref{fig:resolution_bins} empirically proves the necessity of a high-resolution action space. Transitioning from coarse binary routing ($K=2$) to fine-grained bins ($K=10$) nearly doubles average token savings (11.9M $\to$ 20.9M). This confirms that continuous dynamic capacity modulation vastly outperforms rigid High/Low resolution toggles.

\begin{table*}[t] 
\centering
\scriptsize
\setlength{\tabcolsep}{3.5pt} 
\begin{minipage}[t]{0.60\textwidth} 
\vspace{0pt} 
\centering
\caption{\textbf{Router Model Scaling.} MobileNet-Medium with BERT-Tiny (highlighted) offers optimal efficiency-accuracy.}
\label{tab:router_model_size}
\begin{tabular}{l c cc ccc}
\toprule
\textbf{MobileNet} & \textbf{BERT} & \textbf{Acc.} & \textbf{MAE} & \textbf{FLOPs} & \textbf{Lat.} & \textbf{Token Red.} \\
\midrule
Small & Tiny & 72.7 & 1.12 & 12.9 & 437.7 & 50.3\% \\
Small & Mini & 72.8 & 1.13 & 14.4 & 495.9 & 44.1\% \\
Small & Small & 72.6 & 1.12 & 13.1 & 446.2 & 49.4\% \\
Small & Medium & 72.8 & 1.12 & 14.4 & 493.1 & 44.1\% \\
Small & Base & 73.0 & 1.16 & 15.4 & 537.3 & 40.3\% \\
\rowcolor{gray!15} \textbf{Medium} & \textbf{Tiny}  & \textbf{72.8} & \textbf{1.14} & \textbf{14.0} & \textbf{480.8} & \textbf{45.8\%} \\
Medium & Mini & 72.9 & 1.11 & 15.8 & 553.3 & 38.6\% \\
Medium & Small & 72.9 & 1.10 & 15.7 & 549.9 & 38.7\% \\
Medium & Medium & 73.0 & 0.98 & 14.9 & 512.8 & 42.2\% \\
Medium & Base & 72.9 & 1.02 & 14.5 & 498.6 & 43.9\% \\
Large & Tiny & 72.2 & 1.27 & 10.7 & 362.0 & 59.6\% \\
Large & Mini & 73.0 & 1.11 & 16.3 & 582.1 & 36.5\% \\
Large & Small & 72.9 & 1.10 & 14.7 & 505.3 & 42.9\% \\
Large & Medium & 72.9 & 1.04 & 14.5 & 498.9 & 43.6\% \\
Large & Base & 72.7 & 1.08 & 12.2 & 413.4 & 53.2\% \\
\bottomrule
\end{tabular}
\end{minipage}\hfill
\begin{minipage}[t]{0.36\textwidth} 
\vspace{0pt} 
\centering
\caption{\textbf{Oracle Recovery Analysis.} TCRR recovers massive savings in structured tasks but remains conservative in ambiguous scenarios.}
\label{tab:oracle_savings}
\begin{tabular}{l ccc}
\toprule
\textbf{Benchmark} & \textbf{TCRR} & \textbf{Oracle} & \textbf{Recovery \%} \\
\midrule
MMBench & 0.07M & 1.36M & 5.0\% \\
MMStar  & 0.08M & 0.76M & 9.8\% \\
MMMU    & 0.20M & 1.24M & 15.7\% \\
MathVista& 0.34M & 1.25M & 27.2\% \\
Hallusion& 1.02M & 2.83M & 35.9\% \\
OCRBench& 1.91M & 3.22M & 59.3\% \\
MMVet   & 0.25M & 0.77M & 32.3\% \\
TextVQA & 0.10M & 11.46M & 0.9\% \\
ChartQA & 0.01M & 2.92M & 0.3\% \\
DocVQA  & 29.78M & 71.70M & 41.5\% \\
InfoVQA & 25.29M & 34.76M & \textbf{72.8\%} \\
MRWC & 77.41M & 118.16M & \textbf{65.5\%} \\
\bottomrule
\end{tabular}
\end{minipage}
\end{table*}

\subsection{Oracle Analysis and Interpretability (RQ5)}

\textbf{Oracle Gap.} To quantify TCRR's proximity to optimal compression, we define \textit{Recovery (\%)} as actual patch savings divided by the \textit{Oracle Bound} (the theoretical maximum patches safely discarded without semantic degradation, verified via GPT-5.1). Table~\ref{tab:oracle_savings} shows TCRR captures massive savings in structured domains (\textbf{72.8\%} on InfoVQA, \textbf{59.3\%} on OCRBench). Conversely, in ambiguous tasks (e.g., MMBench at $5.0\%$), TCRR adopts a conservative ``safety-first'' policy, preserving resolution to prevent context loss (see routing distributions in Appendix~\ref{app:appendix_stats}).

\textbf{Qualitative Behavior.} TCRR exhibits strictly \textit{query-dependent} routing. Given the exact same image, the router applies aggressive pruning for high-level semantic queries (MMVet) but near-native preservation for detail-heavy instructions (OCR), demonstrating an interpretable, human-aligned decision policy (see visual case studies in Appendix~\ref{app:case_studies}).

\section{Limitations}
\label{sec:limitations}

While TCRR fundamentally optimizes global encoding costs, it currently operates uniformly across the entire image. For ``needle-in-a-haystack'' queries—such as identifying a distant license plate within a 4K street panorama—TCRR must conservatively maintain a high global resolution to preserve the localized detail, yielding minimal upfront FLOPs reduction. However, this boundary highlights a highly complementary relationship: TCRR resolves the dense encoding bottleneck for most scenes, while downstream spatial token pruning (e.g., VisionZip) can seamlessly follow TCRR to discard irrelevant background tokens in these extremely sparse scenarios.    
\section{Conclusion}
\label{sec:conclusion}

In this work, we challenged the prevailing static-resolution paradigm in MLLMs by introducing \textbf{Task-Conditioned Resolution Routing (TCRR)}. By formulating resolution selection as a learnable, query-aware decision process, TCRR systematically decouples visual fidelity from computational cost. Powered by \textbf{Res-500k}—the first large-scale dataset annotated for semantic scale equivalence—TCRR acts as an ultra-lightweight semantic pre-filter. Extensive evaluations demonstrate that TCRR fundamentally shifts the efficiency--accuracy Pareto frontier, cutting FLOPs by up to $41.6\%$ and latency by $48.4\%$ (on Qwen3-VL) with near-zero capability degradation. Our findings underscore a vital design principle: \textit{resolution should no longer be treated as a rigid preprocessing hyperparameter, but as a dynamic, first-class decision variable}. TCRR represents a foundational step toward ``active perception'' in multimodal architectures, paving the way for adaptive systems that tailor their sensory intake strictly to the semantic complexity of the task at hand.



\section*{References}


{
\small
\renewcommand{\bibsection}{} 
\bibliographystyle{unsrtnat} 
\bibliography{reference}   


}


\newpage
\appendix

\section{Data Construction}
\label{app:prompt}
\subsection{GPT Judge Prompt}

\begin{tcolorbox}[
  title={Prompt for Oracle-based Resolution Labeling},
  colback=gray!3,
  colframe=black!40,
  boxrule=0.5pt,
  arc=2pt,
  left=6pt,
  right=6pt,
  top=4pt,
  bottom=4pt
]
\begin{Verbatim}[fontsize=\small, breaklines=true]
You are given a question and multiple answers to the SAME question.

The answer labeled [1.0] represents the correct, complete, and authoritative answer.

Your task is to judge whether other answers lead to the EXACT SAME FINAL RESULT
as the [1.0] answer.

An answer can be considered ACCEPTABLE if and only if:
- It answers the question correctly
- Its final conclusion, result, decision, or key value is the SAME as in [1.0]
- Any differences from [1.0] are purely superficial (wording, style, or brevity)
- No required key information is missing
- No additional incorrect, misleading, or contradictory claims are introduced

An answer should be considered NOT acceptable if:
- The final result is incorrect, incomplete, or ambiguous
- The main conclusion differs in meaning, scope, or implication from [1.0]
- Any key value, constraint, or decision required by the question is wrong or missing
- Extra statements change, weaken, or distort the correct conclusion
- The answer appears correct only due to vague or underspecified wording

IMPORTANT:
- Judge ONLY based on the final outcome and correctness, not on reasoning quality
- Partial correctness is NOT sufficient
- If there is any doubt, mark the answer as NOT acceptable

TASK:
Among answers [0.1] to [0.9], find all answers that are ACCEPTABLE.
From those, select the SMALLEST level.

If none of [0.1,0.2, ... ,0.9] are acceptable, select 1.0.

OUTPUT RULES:
- Output ONE line only
- Do NOT explain
- Do NOT add extra text

Output format:
LOSSLESS_LEVEL=<one of {0.1,0.2,0.3,0.4,0.5,0.6,0.7,0.8,0.9,1.0}>

Question:
{question}

Answers:
[0.1] {answer_0_1}
[0.2] {answer_0_2}
[0.3] {answer_0_3}
[0.4] {answer_0_4}
[0.5] {answer_0_5}
[0.6] {answer_0_6}
[0.7] {answer_0_7}
[0.8] {answer_0_8}
[0.9] {answer_0_9}
[1.0] {answer_1_0}
\end{Verbatim}
\end{tcolorbox}

\subsection{Analysis of Supervision Signals: Why Loss is an Imperfect Proxy}

In the main text (Section~\ref{sec:dataset}), we argued that the \textit{Teacher-Forcing Loss} (or Perplexity Change) utilized in prior works like HyperVL~\cite{hypervl} is a suboptimal supervision signal for resolution routing. Here, we provide a formal analysis of the failure modes inherent to this metric.

Let $\mathcal{L}(I, T, Y_{\text{ref}})$ be the negative log-likelihood (NLL) of the reference response $Y_{\text{ref}}$ given input $(I, T)$:
\begin{equation}
    \mathcal{L}(I, T, Y_{\text{ref}}) = - \sum_{t=1}^{|Y_{\text{ref}}|} \log P_\theta(y_t \mid y_{<t}, I, T).
\end{equation}
Loss-based methods define the optimal scale $s^*_{\text{loss}}$ by thresholding the degradation $\Delta \mathcal{L} = \mathcal{L}(I_s) - \mathcal{L}(I_{\text{orig}})$. We identify two critical regimes where $\Delta \mathcal{L}$ dissociates from true task utility.

\paragraph{Failure Mode 1: The ``Blind Guesser'' (False Efficiency).}
This occurs when the model relies on strong language priors rather than visual evidence.
Consider a query $T$ with a high-probability answer derived from common sense (e.g., ``What color is the sky?'' $\to$ ``Blue''). Even if the image $I_s$ is downsampled to an unrecognizable blur ($s \to 0$), the language model's prior dominates:
\begin{equation}
    P_\theta(Y_{\text{ref}} \mid I_s, T) \approx P_{\text{LM}}(Y_{\text{ref}} \mid T) \approx 1.0.
\end{equation}
In this case, $\Delta \mathcal{L} \approx 0$, leading the algorithm to incorrectly label $s \approx 0$ as sufficient.
\textbf{Consequence:} The router learns to aggressively undersample images for queries with high linguistic predictability, causing hallucinations when the visual reality actually contradicts the prior (e.g., a rare ``orange sky'' at sunset might be compressed to a ``blue sky'' prior).

\paragraph{Failure Mode 2: The ``Lexical Drift'' (False Demand).}
This occurs in open-ended generation tasks (e.g., Captioning) where multiple valid descriptions exist.
Let $Y_{\text{ref}} = \text{``A happy dog runs''}$.
At a compressed scale $I_s$, the stochastic sampling process (or slight feature perturbation) might shift the model's preference to a synonymous caption $Y_s = \text{``A cheerful canine jogs''}$.
Although $Y_s$ is semantically equivalent to $Y_{\text{ref}}$, the token-level cross-entropy is extremely high because the target tokens $y_t$ (``happy'', ``dog'') now have low probability under the distribution conditioned on $I_s$:
\begin{equation}
    P_\theta(\text{``happy''} \mid \text{``A''}, I_s, T) \ll P_\theta(\text{``cheerful''} \mid \text{``A''}, I_s, T).
\end{equation}
Here, $\Delta \mathcal{L} \gg 0$, leading the algorithm to reject $s$ and demand higher resolution.
\textbf{Consequence:} The router learns to preserve resolution merely to maintain \textit{exact lexical matching}, resulting in substantial computational waste for tasks that are semantically robust to compression.

\paragraph{Our Solution: Semantic Equivalence.}
Our proposed metric, $\mathcal{J}(Y_s, Y_{\text{ref}} \mid T)$, operates in the semantic space rather than the token space. By utilizing a powerful judge to assess information retention, we effectively filter out benign lexical drift (solving Mode 2) and enforce factual grounding (solving Mode 1 by checking if visual details like ``text on a sign'' are actually readable in $Y_s$), providing a far cleaner supervision signal for training the router.


\section{Computational Complexity Analysis of TCRR}
\label{app:complexity}

In this section, we provide a rigorous theoretical analysis of TCRR's computational complexity and compare it with empirical latency measurements on NVIDIA H20 GPUs. Our design strictly adheres to a ``lightweight first'' principle, ensuring that the routing overhead remains negligible relative to the backbone inference cost.

\subsection{Input Specifications}
To guarantee constant-time inference regardless of the original raw input size, TCRR operates on fixed-dimension inputs:
\begin{itemize}[leftmargin=*, nosep, itemsep=0pt]
    \item \textbf{Visual Input ($I_{\text{fix}}$):} All images are resized to a standard resolution of $H_{\text{fix}} \times W_{\text{fix}} = 384 \times 384$ pixels.
    \item \textbf{Text Input ($T_{\text{fix}}$):} Text prompts are padded or truncated to a fixed sequence length $L = 192$ tokens. To preserve critical instruction semantics (often at the start) and detailed constraints (often at the end), we employ a \textit{Head-Tail Truncation} strategy. For a prompt with length $L_{\text{raw}} > L$, we construct the input by concatenating the first $\lfloor 0.3 L \rfloor$ tokens and the last $\lceil 0.7 L \rceil$ tokens:
    \begin{equation}
        T_{\text{fix}} = [t_1, \dots, t_{57}] \oplus [t_{L_{\text{raw}}-134}, \dots, t_{L_{\text{raw}}}].
    \end{equation}
\end{itemize}

\subsection{Theoretical FLOPs Decomposition}

We decompose the total floating-point operations (FLOPs) into three primary components: Unimodal Encoding, Cross-Modal Interaction, and the Decision Head.

\paragraph{\textbf{Visual Encoder (MobileNetV4-Medium).}}
The visual backbone serves as the primary computational load, processing $I_{\text{fix}}$ into a feature map $\mathbf{V} \in \mathbb{R}^{C_v \times H' \times W'}$. For MobileNetV4-Medium operating at a resolution of $384^2$, the approximate operation count is:
\begin{equation}
    \mathcal{C}_{\text{visual}} \approx 0.85 \text{ GFLOPs}.
\end{equation}
Although this constitutes the majority of the router's budget, it remains orders of magnitude smaller than standard MLLM visual encoders (e.g., CLIP-ViT-L/14 typically exceeds 80 GFLOPs).

\paragraph{\textbf{Text Encoder (BERT-Tiny).}}
We employ a compact BERT-Tiny architecture (2 layers, hidden dim $D_t=128$, 2 heads). Given the fixed sequence length $L=192$, the complexity is dominated by the linear projections rather than the quadratic attention term:
\begin{align}
    \mathcal{C}_{\text{text}} &\approx 2 L \cdot (4 D_t^2 + 8 D_t^2) + 2 L^2 D_t \nonumber \\
    &\approx 0.012 \text{ GFLOPs}.
\end{align}
The minimal footprint of BERT-Tiny ensures that text processing incurs virtually no latency penalty.

\paragraph{\textbf{Interaction \& Decision Module.}}
This module fuses modalities via FiLM and Cross-Attention. Let $C_v = 512$ denote the visual channel depth and $N = 144$ ($12 \times 12$) be the spatial feature count.
\begin{itemize}[leftmargin=*, nosep, itemsep=0pt]
    \item \textit{FiLM Modulation (2 blocks):} Generating affine parameters $\gamma, \beta$ and applying them:
    \begin{equation}
        \mathcal{C}_{\text{FiLM}} \approx 2 \times [2 D_t C_v + 2 N C_v] \approx 0.0006 \text{ GFLOPs}.
    \end{equation}
    
    \item \textit{Cross-Attention:} The complexity is linear with respect to the number of visual tokens $N$ for the query projection and attention weighting:
    \begin{equation}
        \mathcal{C}_{\text{Cross}} \approx 3 D_h (C_v + D_t) + 2 N D_h \approx 0.0004 \text{ GFLOPs}.
    \end{equation}
    
    \item \textit{Attention Pooling:} Aggregating features via a learnable weight map:
    \begin{equation}
        \mathcal{C}_{\text{Pool}} \approx N C_v + N \approx 0.0001 \text{ GFLOPs}.
    \end{equation}
\end{itemize}

\paragraph{\textbf{Summary.}}
Summing these components, the total theoretical complexity is:
\begin{equation}
    \mathcal{C}_{\text{total}} = \mathcal{C}_{\text{visual}} + \mathcal{C}_{\text{text}} + \mathcal{C}_{\text{interaction}} \approx 0.86 \text{ GFLOPs}.
\end{equation}
This confirms that TCRR consumes approximately $\mathbf{0.8\%}$ of the compute budget required by a typical Qwen3-VL-8B visual encoder ($\sim 100$ GFLOPs), strictly satisfying our design constraint $\mathcal{C}(\pi_\phi) \ll \mathcal{C}(I_{\text{high}})$.

\subsection{Empirical Latency Analysis}

We measure the wall-clock latency on an NVIDIA H20 GPU (FP16 precision, Batch Size=1).

\begin{table}[h]
\centering
\caption{\textbf{TCRR Latency Profiling (NVIDIA H20)}.}
\small
\label{tab:latency_profile}
\begin{tabular}{l c c}
\toprule
\textbf{Component} & \textbf{Latency (ms)} & \textbf{Share (\%)} \\
\midrule
Visual Encoder (MobileNetV4) & 12.4 & 68.5\% \\
Text Encoder (BERT-Tiny) & 2.1 & 11.6\% \\
Interaction Head (FiLM+Attn) & 0.8 & 4.4\% \\
Data Move / Overhead & 2.8 & 15.5\% \\
\midrule
\textbf{Total End-to-End} & \textbf{18.1 ms} & \textbf{100\%} \\
\bottomrule
\end{tabular}
\end{table}

As shown in Table~\ref{tab:latency_profile}, the actual inference takes merely \textbf{18.1 ms}. Even when added to the fastest MLLM backbone execution (e.g., $\sim 250$ ms for Qwen3-VL-2B), the router introduces less than $7\%$ overhead. For larger models like Qwen3-VL-32B ($\sim 2400$ ms), the overhead becomes negligible ($<0.8\%$).

\subsection{Relative Overhead vs. Backbone Savings}

The true viability of TCRR depends on the condition that the routing overhead is strictly marginal compared to the computational savings on the MLLM backbone. 

For a standard ViT within the MLLM, processing $N_v$ patches (where $N_v \propto r / P^2$), the computational complexity of self-attention across $L$ layers is bounded by:
\begin{equation}
    \mathcal{O}_{\text{ViT}}(N_v) = \mathcal{O}(L \cdot N_v \cdot C^2 + L \cdot N_v^2 \cdot C),
\end{equation}
which is quadratic with respect to the input resolution area $r$. 

Conversely, the computational overhead of the TCRR router $\mathcal{O}_{\text{TCRR}}$ is independent of the original high-resolution input $r_{\text{orig}}$. It strictly operates on the proxy image $r_{\text{fix}}$ and comprises the lightweight backbones:
\begin{equation}
    \mathcal{O}_{\text{TCRR}} = \mathcal{O}_{\text{MobileNetV4}}(r_{\text{fix}}) + \mathcal{O}_{\text{BERT}}(L_{\text{text}}) + \mathcal{O}_{\text{Fusion}}.
\end{equation}
Because $r_{\text{fix}}$ is a small constant ($384 \times 384$) and $\mathcal{O}_{\text{Fusion}}$ only consists of low-dimensional MLP and cross-attention operations on a downsampled feature map ($H' \times W'$), $\mathcal{O}_{\text{TCRR}}$ evaluates to approximately $0.86$ GFLOPs (as decomposed above). 

When TCRR routes a query from a high resolution (e.g., yielding $N_{\text{high}}$ tokens) to a lower resolution ($N_{\text{low}}$), the total FLOPs reduction is $\Delta \text{FLOPs} = \mathcal{O}_{\text{ViT}}(N_{\text{high}}) - \mathcal{O}_{\text{ViT}}(N_{\text{low}})$. In our experiments on Qwen3-VL-8B, the savings $\Delta \text{FLOPs}$ often exceed 500 GFLOPs per image, rendering the sub-1 GFLOP router overhead ($<0.2\%$ of the savings) mathematically and practically negligible.


\section{Implementation and Training Details}
\label{app:implementation}

In this section, we provide essential details regarding the data preprocessing pipeline and training protocol to facilitate reproducibility.

\subsection{Input Preprocessing}

\textbf{Visual Padding Strategy (Pad-Keep-Aspect).}
Standard resizing often introduces geometric distortion, which can be detrimental for tasks sensitive to aspect ratios (e.g., OCR). To mitigate this, we employ a \textit{Pad-Keep-Aspect} strategy for the router's input $I_{\text{fix}}$:
\begin{enumerate}
    \item If the raw image dimensions $(H, W)$ are both smaller than the target size $S=384$, the image is placed at the top-left corner of a zero-padded canvas $I_{\text{new}} \in \mathbb{R}^{S \times S}$ without scaling.
    \item Otherwise, the image is scaled down such that its longest side equals $S$ (using Bilinear interpolation) and then centered on the canvas.
\end{enumerate}
This ensures the router perceives the correct geometric structure of the visual content.

\textbf{Target Rescaling Logic.}
As the router operates on a fixed resolution proxy, the regression target must be decoupled from this proxy size. During training, we adjust the ground-truth area scale $s \in (0, 1]$ (relative to the original image) based on the original image's max dimension $M_{\text{orig}}$:
\begin{equation}
    \tilde{s}^* = \begin{cases} 
    s \times \frac{M_{\text{orig}}}{S} & \text{if } M_{\text{orig}} > S \\
    s & \text{otherwise}
    \end{cases}
\end{equation}
This normalization ensures the router learns to predict the optimal downsampling factor relative to the \textit{original} high-resolution input.

\subsection{Training Protocol}

\textbf{Optimization.}
TCRR is trained end-to-end using the Mean Squared Error (MSE) loss. We use the AdamW optimizer with a cosine annealing schedule and a 5\% linear warmup. Training is conducted for 20 epochs, which ensures convergence on the \textbf{Res-500k} dataset. To stabilize training and accelerate throughput, we use Distributed Data Parallel (DDP) with SyncBatchNorm.

\textbf{Hyperparameters.}
Table~\ref{tab:hyperparameters} summarizes the configuration used for our main results.

\begin{table}[h]
\centering
\caption{\textbf{TCRR Training Hyperparameters.}}
\label{tab:hyperparameters}
\begin{tabular}{l c}
\toprule
\textbf{Parameter} & \textbf{Value} \\
\midrule
\multicolumn{2}{l}{\textit{\textbf{Architecture}}} \\
Visual Backbone & MobileNetV4-Medium (e250\_r384) \\
Text Backbone & BERT-Tiny (prajjwal1/bert-tiny) \\
Input Resolution & $384 \times 384$ \\
Max Text Length & 192 \\
MLP Hidden Dim & 512 \\
MLP Dropout & 0.1 \\
\midrule
\multicolumn{2}{l}{\textit{\textbf{Optimization}}} \\
Optimizer & AdamW \\
Learning Rate & $5 \times 10^{-5}$ \\
Weight Decay & $1 \times 10^{-5}$ \\
Batch Size & 32 (per GPU) \\
Epochs & 20 \\
LR Schedule & Cosine Decay \\
Warmup Ratio & 0.05 \\
Seed & 42 \\
\bottomrule
\end{tabular}
\end{table}


\section{Extended Ablation on Fusion Architectures}
\label{app:fusion_architecture}

In Section 6.3 of the main text, we established that our lightweight, single-stage fusion mechanism achieves an optimal balance between routing accuracy and computational overhead. Here, we provide the detailed quantitative comparison supporting this design choice.

As a baseline alternative to our proposed architecture, we designed a \textit{Multi-stage Fusion} variant. In this heavier configuration, the visual and textual features undergo multiple interleaved blocks of cross-attention and feed-forward networks (similar to standard deep multimodal transformer layers), rather than our streamlined, single-pass FiLM and Cross-Attention design.

Table~\ref{tab:fusion_comparison} presents the empirical results measured on the Qwen3-VL backbone. While the multi-stage variant achieves an almost identical token reduction rate ($45.6\%$ vs. $45.8\%$), its increased sequential depth directly penalizes the router's execution time. Specifically, the Router Latency (R. Lat.) increases from $18.1$\,ms to $22.8$\,ms. Because the router acts as an unavoidable pre-filter in the inference pipeline, any delay here directly inflates the end-to-end system latency. Our single-stage TCRR strictly bounds this overhead while achieving the lowest scale prediction error (MAE: $1.14$) and downstream task accuracy ($72.8\%$), validating our ``lightweight first'' architectural philosophy.

\begin{table}[h]
\centering
\caption{\textbf{Router Architecture Comparison.} Comparison of fusion paradigms. The proposed shallow, single-stage fusion achieves the optimal balance of accuracy and computational overhead compared to a heavier multi-stage alternative. ``R. Lat.'' denotes the latency specifically incurred by the router module.}
\label{tab:fusion_comparison}
\small
\setlength{\tabcolsep}{5pt}
\begin{tabular}{l cc cccc}
\toprule
\textbf{Architecture} & \textbf{Acc.} & \textbf{MAE} & \textbf{FLOPs} & \textbf{Lat. (ms)} & \textbf{R. Lat. (ms)} & \textbf{Token Red.} \\
\midrule
Multi-stage Fusion & 72.7 & 1.16 & 14.0 & 499.4 & 22.8 & 45.6\% \\
\textbf{TCRR (Single-stage)} & \textbf{72.8} & \textbf{1.14} & 14.0 & \textbf{480.8} & \textbf{18.1} & \textbf{45.8\%} \\
\bottomrule
\end{tabular}
\end{table}


\section{Additional Statistical Results}
\label{app:appendix_stats}

Table~\ref{tab:resolution_dist} presents the statistical distribution of sample quantities across different resolution intervals (in pixel squared) for each multimodal benchmark dataset, where the resolution intervals are divided into eight levels from $256 \times 256$ to $4096 \times 4096$. This statistical result quantifies the resolution characteristics of diverse multimodal datasets, providing a basis for analyzing the resolution requirements of different multimodal tasks.

\begin{table*}[htbp]
\centering
\caption{\textbf{Resolution Distribution Across Benchmarks.} We classify sample quantities into eight intervals: $R_1: <256^2$, $R_2: 256^2\text{--}384^2$, $R_3: 384^2\text{--}512^2$, $R_4: 512^2\text{--}768^2$, $R_5: 768^2\text{--}1024^2$, $R_6: 1024^2\text{--}2048^2$, $R_7: 2048^2\text{--}4096^2$, and $R_8: >4096^2$ (pixels$^2$). Using indexed headers allows for a cleaner layout and improved readability.}
\label{tab:resolution_dist}
\scriptsize
\setlength{\tabcolsep}{0pt}
\begin{tabular*}{\textwidth}{@{\extracolsep{\fill}} l cccccccc @{}}
\toprule
\textbf{Benchmark} & $R_1$ & $R_2$ & $R_3$ & $R_4$ & $R_5$ & $R_6$ & $R_7$ & $R_8$ \\
\midrule
ChartQA\_TEST         & 8   & 144 & 48   & 2130 & 147 & 23   & 0   & 0   \\
DocVQA\_VAL           & 0   & 0   & 12   & 45   & 196 & 4792 & 257 & 47  \\
HallusionBench        & 14  & 93  & 153  & 262  & 230 & 182  & 11  & 6   \\
InfoVQA\_VAL          & 0   & 0   & 12   & 223  & 368 & 1478 & 630 & 90  \\
MMBench\_DEV\_EN\_V11 & 919 & 974 & 2787 & 12   & 162 & 22   & 0   & 0   \\
MME-RealWorld-CN      & 0   & 0   & 0    & 20   & 141 & 3577 & 1930& 249 \\
MMMU\_DEV\_VAL        & 129 & 222 & 223  & 202  & 122 & 88   & 17  & 0   \\
MMStar                & 197 & 309 & 763  & 147  & 43  & 36   & 5   & 0   \\
MMVet                 & 6   & 15  & 33   & 53   & 42  & 59   & 8   & 2   \\
MathVista\_MINI       & 232 & 200 & 214  & 225  & 58  & 57   & 13  & 1   \\
OCRBench              & 395 & 81  & 101  & 132  & 150 & 115  & 15  & 11  \\
TextVQA\_VAL          & 0   & 0   & 1    & 293  & 4701& 0    & 5   & 0   \\
\bottomrule
\end{tabular*}
\end{table*}

Table~\ref{tab:compression_distribution} shows the distribution of selected compression ratios across benchmarks, revealing strong task-dependent routing patterns.

\begin{table*}[t]
\centering
\caption{Distribution of selected compression ratios across benchmarks.}
\label{tab:compression_distribution}
\scriptsize
\setlength{\tabcolsep}{0pt}
\begin{tabular*}{\textwidth}{@{\extracolsep{\fill}}l cccccccccc @{}}
\toprule
\textbf{Dataset} & \textbf{0.1} & \textbf{0.2} & \textbf{0.3} & \textbf{0.4} & \textbf{0.5} & \textbf{0.6} & \textbf{0.7} & \textbf{0.8} & \textbf{0.9} & \textbf{1.0} \\
\midrule
ChartQA\_TEST            & 314          & 734          & 626          & 239          & 114          & 57           & 61           & 25           & 24           & 134          \\
DocVQA\_VAL              & 4356         & 599          & 169          & 84           & 37           & 20           & 19           & 11           & 9            & 33           \\
HallusionBench           & 632          & 53           & 33           & 11           & 5            & 5            & 1            & 2            & 1            & 5            \\
InfoVQA\_VAL             & 1325         & 505          & 337          & 202          & 116          & 55           & 56           & 36           & 27           & 123          \\
MathVista\_MINI          & 303          & 79           & 41           & 26           & 8            & 12           & 7            & 5            & 4            & 11           \\
MMBench\_DEV\_EN\_V11    & 930          & 75           & 37           & 15           & 8            & 11           & 3            & 5            & 10           & 14           \\
MMMU\_DEV\_VAL           & 300          & 33           & 14           & 10           & 8            & 7            & 8            & 3            & 2            & 46           \\
MMStar                   & 328          & 68           & 38           & 19           & 5            & 10           & 1            & 3            & 11           & 25           \\
MMVet                    & 100          & 27           & 9            & 4            & 6            & 4            & 6            & 2            & 5            & 15           \\
OCRBench                 & 200          & 93           & 65           & 36           & 22           & 10           & 8            & 4            & 4            & 16           \\
TextVQA\_VAL             & 2754         & 873          & 452          & 249          & 159          & 110          & 85           & 54           & 43           & 204          \\
MME-RealWorld-CN         & 4150         & 765          & 365          & 216          & 129          & 85           & 51           & 42           & 28           & 82           \\
\bottomrule
\end{tabular*}
\end{table*}


\section{TCRR: Qualitative Case Studies}
\label{app:case_studies}

Below we provide comprehensive visual case studies to explicitly demonstrate how \textbf{Task-Conditioned Resolution Routing (TCRR)} dynamically assigns resolution area scaling ratios based on the semantic demands of different textual prompts. This mechanism aims to find the optimal Pareto balance between feature retention and computational efficiency, rather than blindly pursuing maximum compression.

\subsection{Category 1: Intra-Image Prompt Sensitivity (Same Image, Different Routing)}
This category demonstrates TCRR's core novelty: resolution allocation is dynamically determined by the text prompt, not just the visual complexity of the image itself. 

\subsubsection{Case 1A: 4K Urban Landscape}
\textbf{Original Resolution:} $3840 \times 2523$ pixels.

\begin{table}[h!]
    \centering
    \small
    \renewcommand{\arraystretch}{1.5}
    \renewcommand{\tabularxcolumn}[1]{m{#1}} 
    \begin{tabularx}{\textwidth}{@{} 
        >{\raggedright\arraybackslash}m{0.14\textwidth} 
        >{\raggedright\arraybackslash}X 
        >{\raggedright\arraybackslash}m{0.16\textwidth} 
        >{\centering\arraybackslash}m{0.16\textwidth} 
        >{\raggedright\arraybackslash}X 
    @{}}
    \toprule
    \textbf{Task Type} & \textbf{Prompt} & \textbf{TCRR Routing} & \textbf{Visual Input} & \textbf{Analysis} \\
    \midrule
    \textbf{Global Semantics} & 
    \textit{``Is this picture taken during the day or at night?''} & 
    \textbf{Area Scale: 10\%}\newline \scriptsize{(Eq. Res: $\sim \! 1214 \!\times\! 798$)} & 
    \includegraphics[width=0.15\textwidth]{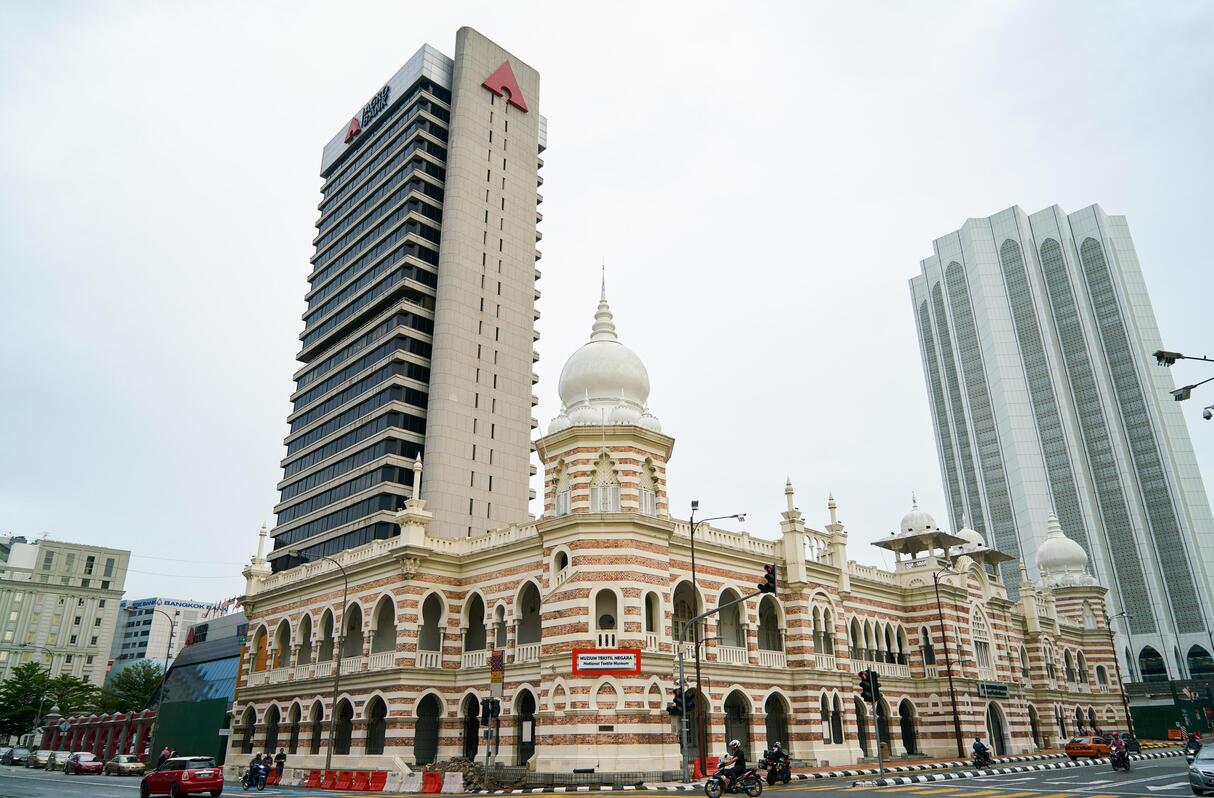} & 
    While pruning \textbf{90\%} of visual tokens, the model retains sufficient global illumination features to correctly output \textbf{``Daytime''}. This is the optimal computational sweet spot for high-confidence predictions. \\
    \midrule
    \textbf{Fine Details} & 
    \textit{``What is the text inside the red rectangular area on the building in the bottom center?''} & 
    \textbf{Area Scale: 100\%}\newline \scriptsize{(Orig. Res: $3840 \!\times\! 2523$)} & 
    \includegraphics[width=0.15\textwidth]{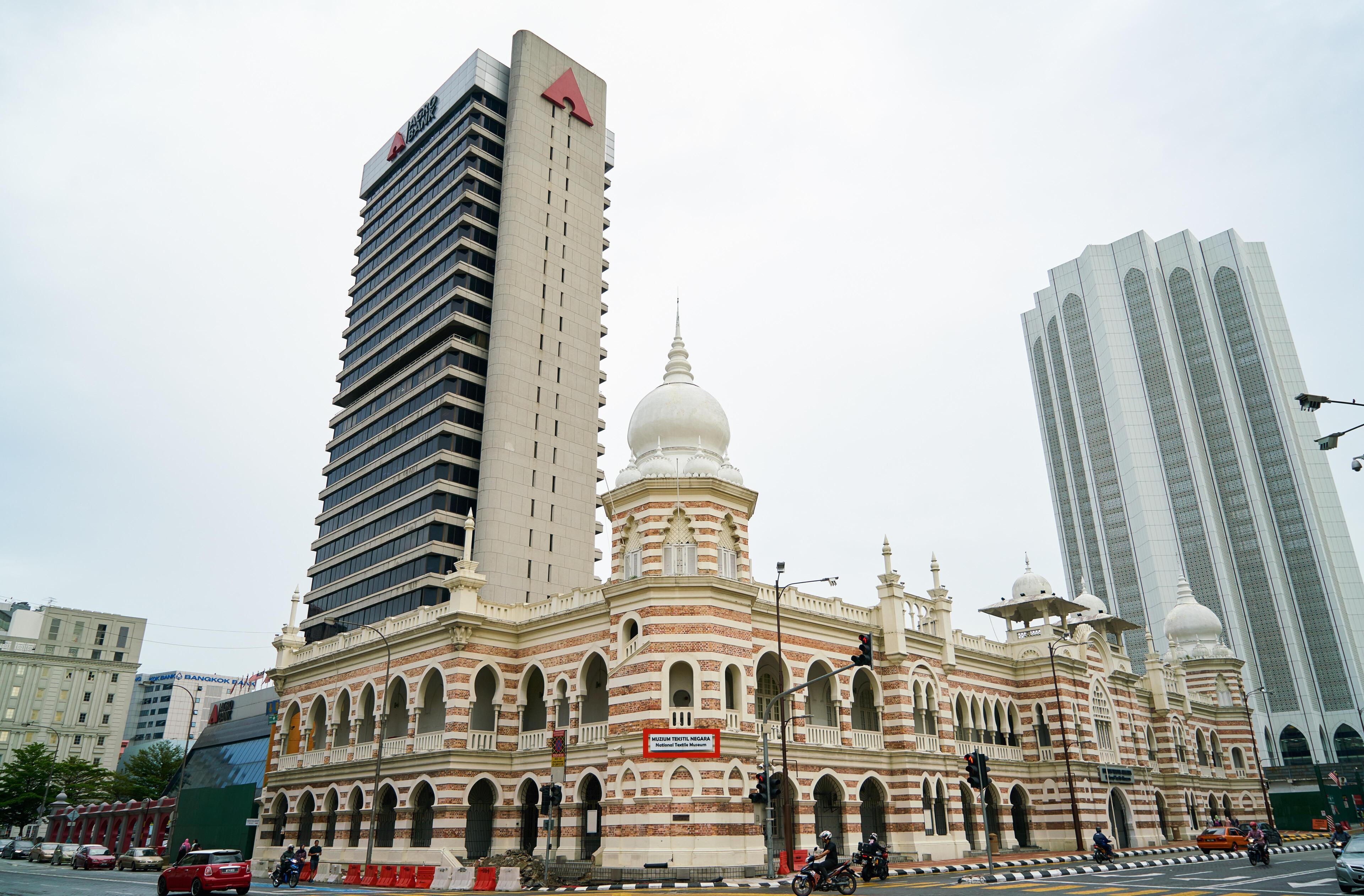} & 
    The model dynamically determines that pixel-level precision is required. It bypasses down-sampling to accurately read: \textbf{``MUZIUM TEKSTIL NEGARA / National Textile Museum''}. \\
    \bottomrule
    \end{tabularx}
\end{table}

\vspace{1em}
\subsubsection{Case 1B: 2K Indoor Study Area}
\textbf{Original Resolution:} $2048 \times 2048$ pixels.

\begin{table}[h!]
    \centering
    \small
    \renewcommand{\arraystretch}{1.5}
    \renewcommand{\tabularxcolumn}[1]{m{#1}} 
    \begin{tabularx}{\textwidth}{@{} 
        >{\raggedright\arraybackslash}m{0.14\textwidth} 
        >{\raggedright\arraybackslash}X 
        >{\raggedright\arraybackslash}m{0.16\textwidth} 
        >{\centering\arraybackslash}m{0.16\textwidth} 
        >{\raggedright\arraybackslash}X 
    @{}}
    \toprule
    \textbf{Task Type} & \textbf{Prompt} & \textbf{TCRR Routing} & \textbf{Visual Input} & \textbf{Analysis} \\
    \midrule
    \textbf{Global Semantics} & 
    \textit{``What is the functional scene of this room?''} & 
    \textbf{Area Scale: 10\%}\newline \scriptsize{(Eq. Res: $\sim \! 648 \!\times\! 648$)} & 
    \includegraphics[width=0.15\textwidth]{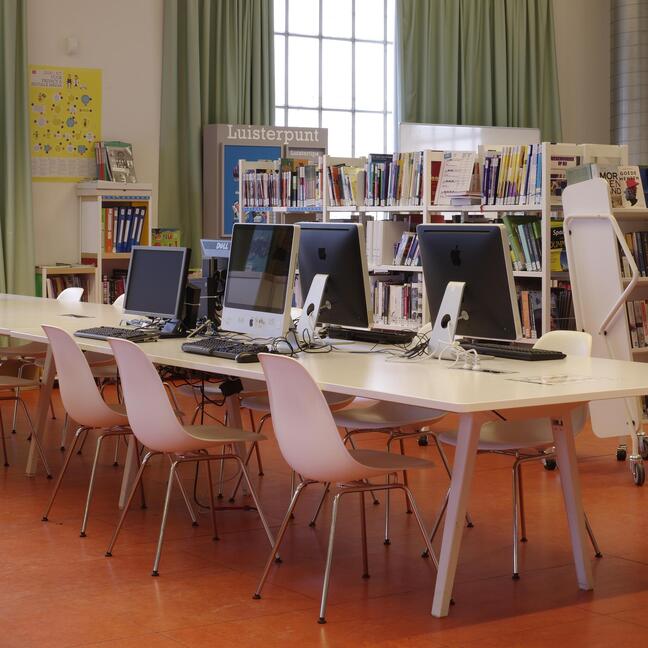} & 
    After aggressively removing \textbf{90\%} of local redundancy, the model relies on core structural features to correctly output: \textbf{``Computer lab or library reading area''}. \\
    \midrule
    \textbf{Fine Details} & 
    \textit{``What brands are the computers on the desk?''} & 
    \textbf{Area Scale: 90\%}\newline \scriptsize{(Eq. Res: $\sim \! 1943 \!\times\! 1943$)} & 
    \includegraphics[width=0.15\textwidth]{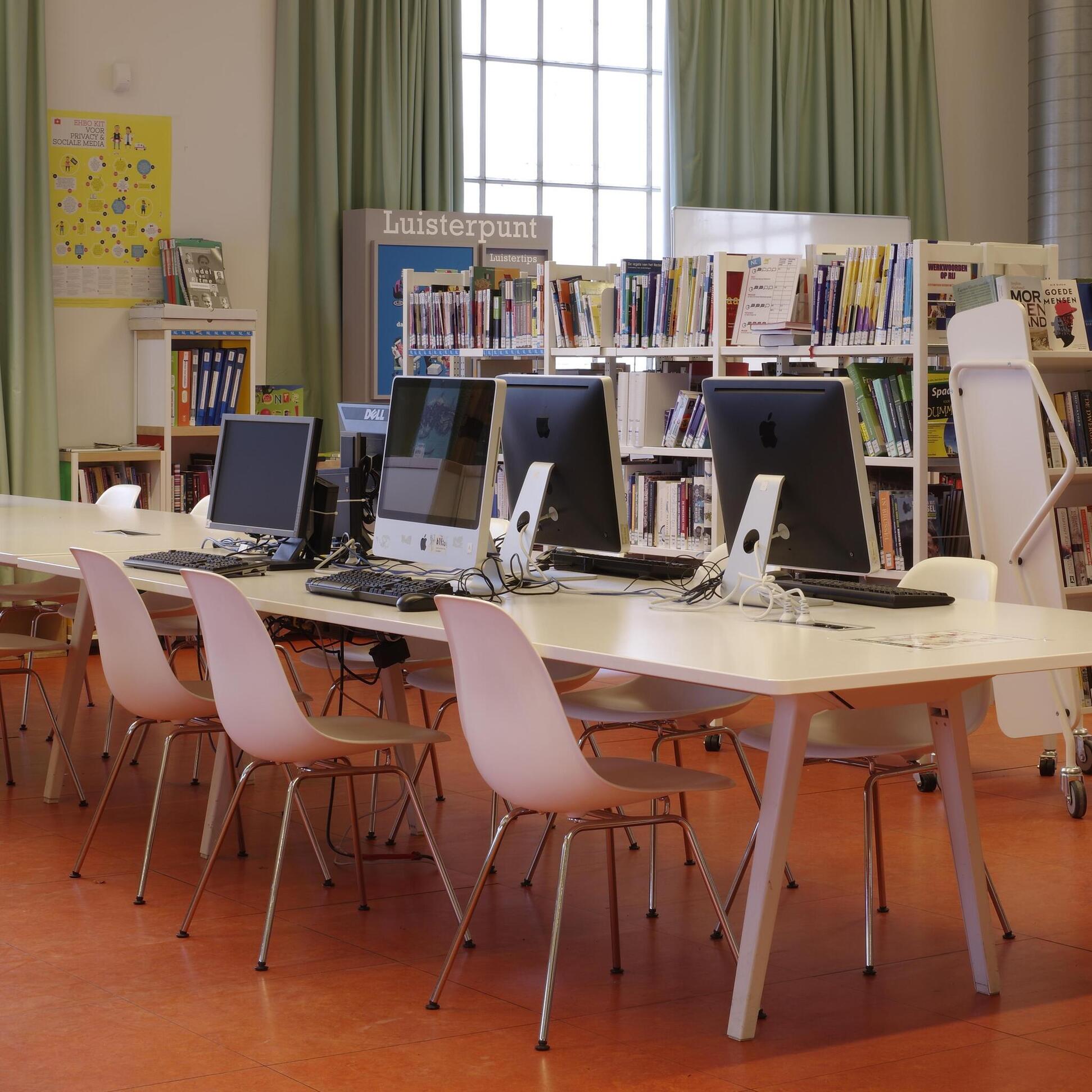} & 
    TCRR intelligently assigns a high resolution to extract micro-features (the tiny logo on the back of the distant Dell monitor would blur entirely at low resolutions). Output: \textbf{``Apple and Dell''}. \\
    \bottomrule
    \end{tabularx}
\end{table}

\clearpage 
\subsection{Category 2: High-Resolution Retention for Dense Information}
For images with dense text or complex structures (e.g., charts, documents), TCRR conservatively maintains high pixel retention rates to prevent the loss of critical fine-grained structures during global compression.

\subsubsection{Case 2A: Chart QA (ChartQA)}
\textbf{Original Resolution:} $800 \times 836$ pixels.

\begin{table}[h!]
    \centering
    \small
    \renewcommand{\arraystretch}{1.5}
    \renewcommand{\tabularxcolumn}[1]{m{#1}} 
    \begin{tabularx}{\textwidth}{@{} 
        >{\raggedright\arraybackslash}m{0.14\textwidth} 
        >{\raggedright\arraybackslash}X 
        >{\raggedright\arraybackslash}m{0.16\textwidth} 
        >{\centering\arraybackslash}m{0.16\textwidth} 
        >{\raggedright\arraybackslash}X 
    @{}}
    \toprule
    \textbf{Task Type} & \textbf{Prompt} & \textbf{TCRR Routing} & \textbf{Visual Input} & \textbf{Analysis} \\
    \midrule
    \textbf{Information Extraction} & 
    \textit{``Which airport was the most used?''} & 
    \textbf{Area Scale: 80\%}\newline \scriptsize{(Eq. Res: $\sim \! 716 \!\times\! 748$)} & 
    \includegraphics[width=0.15\textwidth]{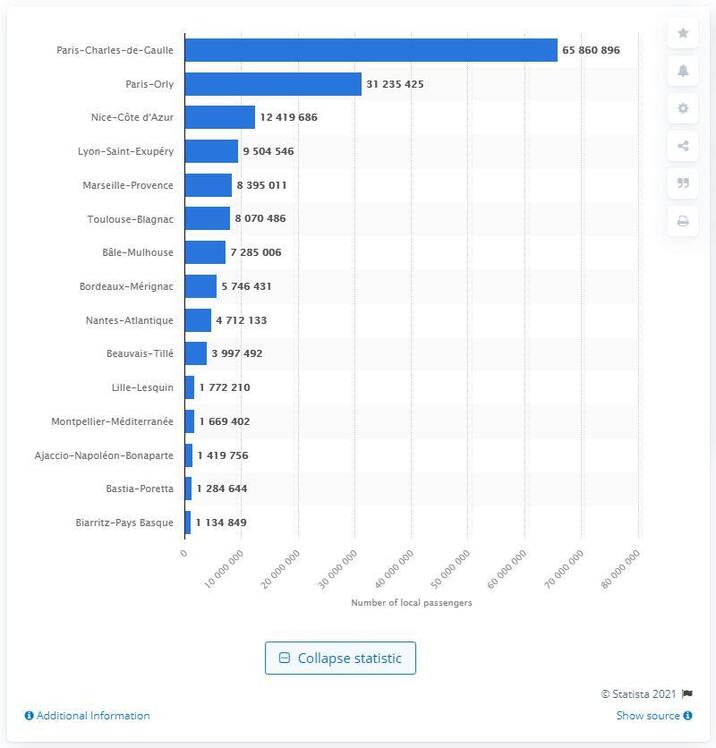} & 
    Recognizing dense axis texts and fine-grained data bars, TCRR adopts an extremely conservative down-sampling strategy. It marginally reduces resolution to save 20\% compute while preserving impeccable legibility, correctly outputting: \textbf{``Paris-Charles-de-Gaulle''}. \\
    \bottomrule
    \end{tabularx}
\end{table}

\vspace{1em}
\subsubsection{Case 2B: Dense Document QA (DocVQA)}
\textbf{Original Resolution:} $1679 \times 2340$ pixels.

\begin{table}[h!]
    \centering
    \small
    \renewcommand{\arraystretch}{1.5}
    \renewcommand{\tabularxcolumn}[1]{m{#1}} 
    \begin{tabularx}{\textwidth}{@{} 
        >{\raggedright\arraybackslash}m{0.14\textwidth} 
        >{\raggedright\arraybackslash}X 
        >{\raggedright\arraybackslash}m{0.16\textwidth} 
        >{\centering\arraybackslash}m{0.16\textwidth} 
        >{\raggedright\arraybackslash}X 
    @{}}
    \toprule
    \textbf{Task Type} & \textbf{Prompt} & \textbf{TCRR Routing} & \textbf{Visual Input} & \textbf{Analysis} \\
    \midrule
    \textbf{Document Parsing} & 
    \textit{``What is the \% of raw material imported in the current year?''} & 
    \textbf{Area Scale: 90\%}\newline \scriptsize{(Eq. Res: $\sim \! 1593 \!\times\! 2220$)} & 
    \includegraphics[width=0.15\textwidth]{figures/Case_2B_1593_2220.jpg} & 
    To extract a specific percentage from a financially dense table, TCRR assigns near-maximum resolution (90\%). This strictly guarantees that microscopic digits do not suffer from aliasing before entering the ViT, enabling the model to precisely extract: \textbf{``79.23\%''}. \\
    \bottomrule
    \end{tabularx}
\end{table}

\clearpage
\subsection{Category 3: Extreme Redundancy Pruning}
For high-level semantic queries (e.g., mood perception or basic object recognition), fine-grained details are computationally inefficient. TCRR aggressively scales down the entire image (pruning 80\%--90\% of pixel area) to extract core semantics at ultra-low latency.

\subsubsection{Case 3A: Landscape (Mood Perception)}
\textbf{Original Resolution:} $580 \times 381$ pixels.

\begin{table}[h!]
    \centering
    \small
    \renewcommand{\arraystretch}{1.5}
    \renewcommand{\tabularxcolumn}[1]{m{#1}} 
    \begin{tabularx}{\textwidth}{@{} 
        >{\raggedright\arraybackslash}m{0.14\textwidth} 
        >{\raggedright\arraybackslash}X 
        >{\raggedright\arraybackslash}m{0.16\textwidth} 
        >{\centering\arraybackslash}m{0.16\textwidth} 
        >{\raggedright\arraybackslash}X 
    @{}}
    \toprule
    \textbf{Task Type} & \textbf{Prompt} & \textbf{TCRR Routing} & \textbf{Visual Input} & \textbf{Analysis} \\
    \midrule
    \textbf{Mood Perception} & 
    \textit{``What is the overall mood or atmosphere conveyed by this landscape?''} & 
    \textbf{Area Scale: 20\%}\newline \scriptsize{(Eq. Res: $\sim \! 259 \!\times\! 170$)} & 
    \includegraphics[width=0.15\textwidth]{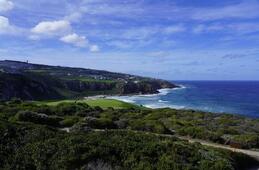} & 
    Pixel-level textures (individual leaves, tiny ripples) are redundant for mood perception. TCRR aggressively scales the image down to 20\% to capture global color blocks and contours, accurately predicting: \textbf{``Serene and natural''} at ultra-low latency. \\
    \bottomrule
    \end{tabularx}
\end{table}

\vspace{1em}
\subsubsection{Case 3B: Object Recognition (Salient Subject)}
\textbf{Original Resolution:} $680 \times 454$ pixels.

\begin{table}[h!]
    \centering
    \small
    \renewcommand{\arraystretch}{1.5}
    \renewcommand{\tabularxcolumn}[1]{m{#1}} 
    \begin{tabularx}{\textwidth}{@{} 
        >{\raggedright\arraybackslash}m{0.14\textwidth} 
        >{\raggedright\arraybackslash}X 
        >{\raggedright\arraybackslash}m{0.16\textwidth} 
        >{\centering\arraybackslash}m{0.16\textwidth} 
        >{\raggedright\arraybackslash}X 
    @{}}
    \toprule
    \textbf{Task Type} & \textbf{Prompt} & \textbf{TCRR Routing} & \textbf{Visual Input} & \textbf{Analysis} \\
    \midrule
    \textbf{Basic Recognition} & 
    \textit{``What is the main animal lying right in the center?''} & 
    \textbf{Area Scale: 10\%}\newline \scriptsize{(Eq. Res: $\sim \! 215 \!\times\! 144$)} & 
    \includegraphics[width=0.15\textwidth]{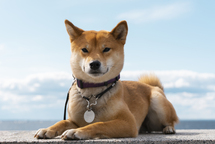} & 
    Iconic global features of a Shiba Inu (reddish-gold coat, triangular ears, silhouette) remain distinctly recognizable even when uniformly compressed to 10\% of the original area. TCRR assigns the lowest resolution routing, maximizing efficiency while fully preserving semantic accuracy: \textbf{``Shiba Inu''}. \\
    \bottomrule
    \end{tabularx}
\end{table} 




\end{document}